\def\year{2021}\relax
\documentclass[letterpaper]{article} 
\usepackage{aaai21}  
\usepackage{times}  
\usepackage{helvet} 
\usepackage{courier}  
\usepackage[hyphens]{url}  
\usepackage{graphicx} 
\usepackage{graphicx}  
\usepackage{natbib}  
\usepackage{caption} 
\usepackage{graphicx}
\usepackage{comment}
\usepackage{amsmath,amssymb} 
\usepackage{color}

\DeclareMathOperator*{\argmax}{argmax}
\DeclareMathOperator*{\argmin}{argmin}
\usepackage{url}
\usepackage{booktabs}
\usepackage{nicefrac}
\usepackage{microtype}
\usepackage{epsfig}
\usepackage{subcaption}
\usepackage{amsmath}
\usepackage{amssymb}
\usepackage{amsthm}
\usepackage{bm}
\usepackage{multirow}
\usepackage{makecell}
\usepackage{footmisc}
\usepackage{marvosym}
\usepackage{latexsym}
\usepackage[utf8]{inputenc}
\usepackage{kotex}
\usepackage{titlesec}
\usepackage{caption}
\definecolor{mypink2}{RGB}{219, 48, 122}

\newcommand{\ie}{\textit{i}.\textit{e}., }
\newcommand{\eg}{\textit{e}.\textit{g}., } 
\newcommand{\et}{\textit{et al.}}

\title{Associative Partial Domain Adaptation}

\makeatletter
\newcommand{\printfnsymbol}[1]{%
  \textsuperscript{\@fnsymbol{#1}}%
}
\makeatother

\author{Youngeun Kim\textsuperscript{\rm 1} \thanks{Equal contribution},
Sungeun Hong\textsuperscript{\rm 2} \printfnsymbol{1},
Seunghan Yang\textsuperscript{\rm 3}, Sungil Kang\textsuperscript{\rm 2}, Yunho Jeon\textsuperscript{\rm 4}, Jiwon Kim\textsuperscript{\rm 2} \\
\small{
\textnormal{
\textsuperscript{\rm 1}Yale University, USA\thanks{Work done while being at SK Telecom}\:\:
\textsuperscript{\rm 2}SK telecom, South Koea \:\:
\textsuperscript{\rm 3}KAIST, South Korea \:\:
\textsuperscript{\rm 4}mofl Inc., South Korea\printfnsymbol{2} \\
youngeun.kim@yale.edu, csehong@gmail.com, seunghan@kaist.ac.kr, sung1.kang@sk.com, jyh2986@gmail.com, kjw0612@gmail.com
}}}

\begin{document}

\maketitle

\begin{abstract}
Partial  Adaptation (PDA) addresses a practical scenario in which the target domain contains only a subset of classes in the source domain. While PDA should take into account both class-level and sample-level to mitigate negative transfer, current approaches mostly rely on only one of them. 
In this paper, we propose a novel approach to fully exploit multi-level associations that can arise in PDA. 
Our Associative Partial Domain Adaptation (APDA) utilizes \textit{intra-domain association} to actively select out non-trivial anomaly samples in each source-private class that sample-level weighting cannot handle. 
Additionally, our method considers \textit{inter-domain association} to encourage positive transfer by mapping between nearby target samples and source samples with high label-commonness.
For this, we exploit feature propagation in a proposed label space consisting of source ground-truth labels and target probabilistic labels.
We further propose a geometric guidance loss based on the label commonness of each source class to encourage positive transfer.
Our APDA consistently achieves state-of-the-art performance across public datasets.
\end{abstract}

\section{Introduction}
\label{sec:intro}

In general, existing domain adaptation approaches ~\cite{saenko2010adapting,tzeng2014deep,ganin2015unsupervised} assume that the identical label set is shared between the source and target domains.
However, finding a source domain with the same label set as the target domain of interest is very difficult and burdensome in the real-world \cite{hong2017sspp,hoffman2018cycada}.
To alleviate the constraint of using a shared identical label set in domain adaptation, Partial Domain Adaptation (PDA) was first introduced by Cao \et~\cite{cao2018partialtransfer}.
In PDA, unlike standard domain adaptation, the target domain only has a subset of the source labels.
The target label information cannot be accessed during training, and thus the size of the target label set is unknown~\cite{li2020deep}.
Therefore, simply applying existing domain adaptation approaches to PDA can result in negative transfer due to source-private classes irrelevant to the target domain \cite{cao2018partialtransfer}.

\begin{figure}[t]
\begin{center}
\def\arraystretch{0.5}
\begin{tabular}{@{}c@{\hskip 0.02\linewidth}c@{\hskip 0.02\linewidth}c}
\includegraphics[width=0.32\linewidth]{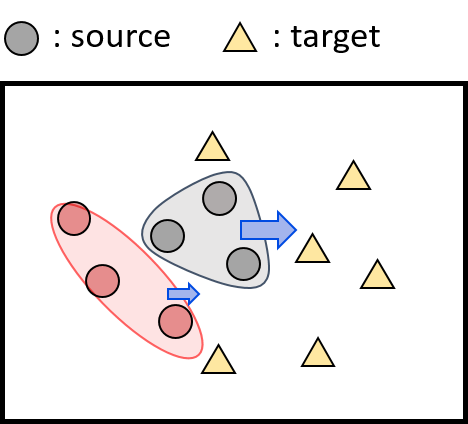} &
\includegraphics[width=0.32\linewidth]{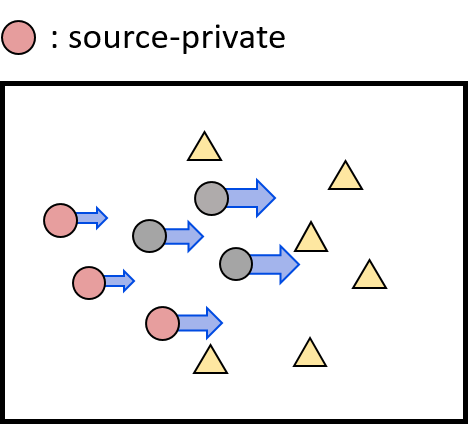}  &
\includegraphics[width=0.32\linewidth]{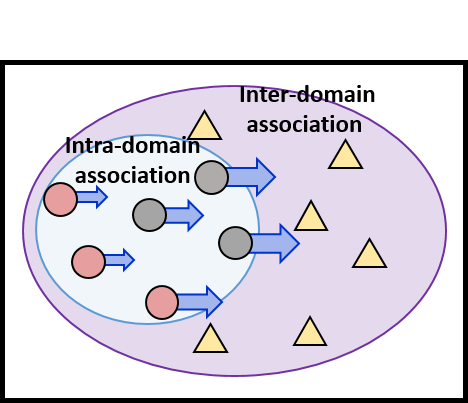} 
\\
\vspace*{-0.1in}
{(a)} & {(b)} &  {(c)}  \\
\end{tabular}
\end{center}

\caption{
Comparison of the PDA approaches. The arrows indicate the direction and magnitude of the source sample's movement by domain confusion loss. 
 (a) Class-level weighting gives equal weight to samples in the same class without considering the characteristics of each sample.
 (b) Sample-level weighting enables better PDA via measuring the commonness of each sample depending on its location.
 Unfortunately, this approach still suffers from negative transfer due to inter-class variation.
 (c) Our multi-level associative weighting can effectively handle anomaly samples,  taking into account intra- and inter-domain class relationships.}
\vspace*{-0.1in}
\label{fig:motivation}
\end{figure}

The main technical challenge in PDA is to isolate the source-private classes during the domain adaptation process; and thus, PDA necessarily entails estimating label-commonness, a measure of whether each source class or sample overlaps with target domain labels.
Early PDA approaches~\cite{cao2018partialtransfer,cao2018partial} exploit class-level weighting by observing that the output of the source classifier for the target samples reflects the distribution of the source label space.
Concretely, they obtain class-specific weights by averaging label predictions (\ie class probabilities) across the target dataset and then apply them when training the label classifier and adversarial domain discriminator.
However, this approach could be vulnerable to the class imbalance problem of the target dataset due to its high dependency on the whole target domain samples as mentioned in  \cite{zhang2018importance}. 
Moreover, giving the same weight to all samples within the same class does not fully utilize the characteristics of each sample as shown in Fig \ref{fig:motivation} (a).

Recent state-of-the-art methods commonly exploit sample-level weighting mechanisms \cite{zhang2018importance,cao2019learning}.
Despite the recent promising results, there are several issues not addressed well in the existing sample-level approaches~\cite{zhang2018importance,cao2019learning}.
First, they still suffer from negative transfer from large intra-class variation within each source-private class.
For example, due to non-trivial intra-class variation, several source-private samples can be located near the target samples, and some of the source samples with common labels can be far from the target samples as shown in Fig. \ref{fig:motivation} (b).
As a result, it is difficult to measure commonness only at the sample-level due to unclear class boundaries and especially the presence of anomaly samples.
Second, they do not explicitly consider inter-dependency between source and target domains while seeking label-commonness.
We argue that if the model can learn a structure that reflects the intra- and inter-domain relationship as shown in Fig \ref{fig:motivation} (c), the performance can be improved further.

In this paper, we propose  Associative Partial Domain Adaptation (APDA) that explicitly takes into account intra- and inter-domain associations by leveraging feature propagation in the label space.
The proposed \textit{intra-domain association} can learn an intrinsic structure within each class, which facilitates to select out anomaly samples in each source-private class.
Also, source and target samples with high label-commonness are encouraged to be located nearby in the feature space by \textit{inter-domain association}, and thus positive transfer is enhanced. 
To this end, we explicitly learn the inter-dependent structure between the source and target domains by graph-based propagation.
One of the important issues when applying graph-based propagation to the non-structural input is the definition of adjacency matrices (\ie edges).
Our proposed APDA exploits the similarity of class labels to construct edges.
Specifically, we use hard labels in the source domain and predicted soft labels (\ie probabilistic labels) in the target domain.
Furthermore, we introduce a novel confidence-guided loss based on the moving-averaged commonness score of source samples.  
The key idea of our confidence-guided loss is to keep the source-private classes away from the target sample while keeping the common classes close to the target samples in the feature space.

The major contributions of this study are as follows: 
(1) Our associative feature propagation can prevent negative transfer due to non-trivial intra-class variation and can explicitly learn the structure of intra- and inter-domain relationships.
(2) We propose a novel guidance loss that enhances positive transfer and suppresses negative transfer via label-commonness.
(3) On public PDA benchmarks, our model consistently achieves state-of-the-art performance. 
 Also, we quantitatively and qualitatively demonstrate how our associative modeling efficiently solves PDA problems.

\section{Related Work}
\label{sec:related_work}
\subsection{Domain Adaptation}
\label{ssec:da}

Early investigations on deep domain adaptation involve Maximum Mean Discrepancy (MMD)~\cite{tzeng2014deep,long2015learning} that enables features from two different domains to resemble each other.
Long \et~\cite{long2016unsupervised} exploit residual transfer modules that can bridge the source classifier and separate target classifier.
Taking into account the explicit matching of higher-order moments, Central Moment Discrepancy (CMD) \cite{sun2016deep} and second-order statistics matching~\cite{sun2016deep} have been proposed.
Meanwhile, domain adversarial training \cite{ganin2015unsupervised,hong2019unsupervised,chang2019domain,hong2020attention,choi2020hi,kim2020domain} has also been extensively studied.
They commonly use adversarial deep neural networks where the label classifier trained from the labeled source domain aims to generate discriminative features while the adversarial domain discriminator makes the features domain-invariant.
Recently, a technique using graph convolution \cite{kipf2016semi} has been proposed, but it has not been actively studied so far \cite{ma2019gcan}.
However, crucially, existing domain adaptation methods assume that the source and target domains share the same label space, which is a strict constraint in the real world.

\subsection{Partial Domain Adaptation}
\label{ssec:pda}
In an effort to alleviate the constraint of using identical label space in domain adaptation, Partial Domain Adaptation (PDA) \cite{cao2018partialtransfer,cao2018partial} addresses the scenario in which the source label space is large enough to fully cover the target label space.
The main technical challenge of PDA is negative transfer caused by source-private classes.
Selective Adversarial Networks (SAN)~\cite{cao2018partialtransfer} adopt multiple domain discriminators for each class and suggest weighting mechanisms based on class probabilities to select out source-private classes.
Cao \et\cite{cao2018partial} further introduces a simple yet effective way of using only one adversarial domain discriminator.

Recently, Zhang \et\cite{zhang2018importance} suggest a sample-level weighting mechanism based on the activation of the auxiliary domain discriminator.
They insist that existing class-level weighting  \cite{cao2018partialtransfer,cao2018partial} cannot cope with class imbalanced target domains due to the process of averaging the class probabilities across the target domain.
Example Transfer Network (ETN) \cite{cao2019learning} further improves the quality of feature transferability by integrating the discriminative information into the sample-level weighting mechanism.
Overall, the sample-level weighting has the advantage of being less vulnerable to target class imbalance~\cite{zhang2018importance}, but it has difficulty handling large intra-class variability within each source-private class.
We solve this issue by our associative mechanism in a joint label space with the hard labels (from the source domain) and the probabilistic labels (from the target domain).

\begin{figure*}
     \centering
         \includegraphics[width=0.75\textwidth]{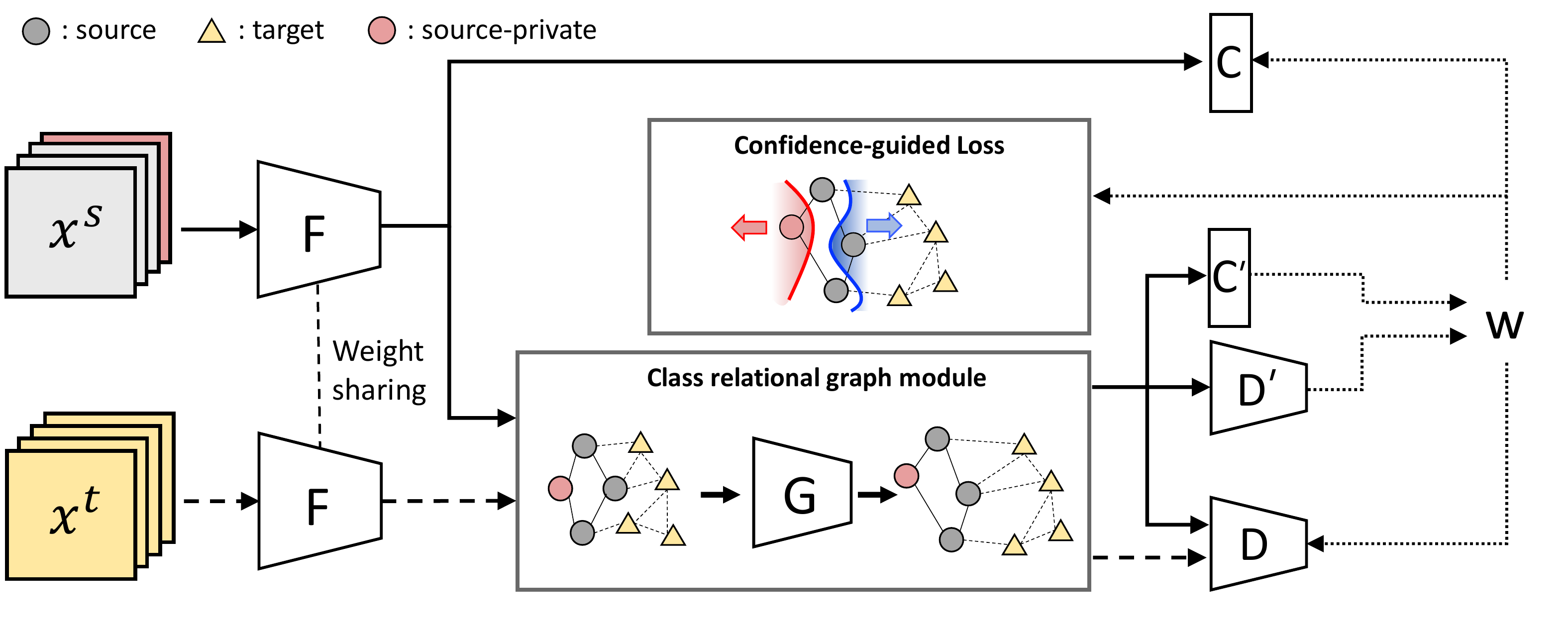}
 \caption{
 Overall flow of APDA framework. 
 (1)  For clarity, we start with a sample-level weighting mechanism based on the label prediction uncertainty (from $C'$) and domain prediction  (from $D'$) of each source sample.
 We then adjust the influence of each sample according to the label commonness in a label classifier $C$ and an adversarial domain discriminator $D$.
 (2) However, sample-level weighting alone is not enough to mitigate the negative transfer caused by intra-class variation. Thus, we introduce class relational graph in which samples with similar labels communicate with each other.
 (3) Finally, we use a confidence-guided loss that keeps the source-private classes away from the target samples while keeping the common classes close to the target samples.
 Notice that all the training processes (1) to (3) are conducted simultaneously.
 }
     \label{fig:Main_pipeline}
\end{figure*}


\section{Method}
\label{sec:method}
The main goal of Partial Domain Adaptation (PDA) is to minimize  discrepancy between a labeled source domain  $D_{s}=\{\left(\textbf{x}^{i}_{s}, y^{i}_{s}\right)\}^{n_{s}}_{i=1}$ and a unlabeled target domain $D_{t}=\{\textbf{x}^{j}_{t}\}^{n_t}_{{j=1}}$.
As in the standard DA, PDA assumes that source and target samples are drawn from a different but related probability distributions $D_{s} \sim p_{s}$ and $D_{t} \sim p_{t}$, respectively.
The main difference between PDA and the standard DA is that source class label set subsumes target label set in PDA, \ie $\mathbb{C}_{s} \supset \mathbb{C}_{t}$.
Note that any target label information, such as the total number of classes, cannot be accessed during training, and thus simply applying common domain adaptation techniques to PDA can result in negative transfer.

\subsection{Background and Motivation}
\label{ssec:bg_motivation}

The main assumption of domain adversarial training is that if the learned features are domain-invariant, a classifier trained on the labeled source domain can perform well even for the unlabeled target domain.
Domain adversarial networks consist of three main modules: feature extractor $F(\cdot|\theta _{f})$, label classifier $C(\cdot|\theta _{c})$, and  domain discriminator $D(\cdot|\theta _{d})$, where $\theta _{f}$, $\theta _{c}$, and $\theta _{d}$ denote the parameter of each part. 
The objective of the label classifier is to predict the category for the labeled source domain.
Importantly, the domain discriminator attempts to distinguish the domain of the samples, while the feature extractor tries to fool the domain discriminator, resulting in domain-invariant features.
The overall training objective can be formulated as follows:
\begin{equation}
    \label{eq:minmax}
    \begin{split}
L(\theta _{f}, \theta _{c}, \theta _{d}) &= \mathbb{E}_{(\mathbf{x}_{s},y_s)\sim p_{s}}[L_{y}(C(F(\mathbf{x}_{s})), y_s)] \\ 
& - \mathbb{E}_{(\mathbf{x}, d)\sim p}[L_{d}(D(F(\mathbf{x})), d)],
\end{split}
\end{equation}
where $p$ refers the sample distribution, $d$ is the domain label, and $L_{y}(\cdot)$ and $L_{d}(\cdot)$ denote the cross-entropy loss function respectively.
The parameters ${\theta}_f, {\theta}_c$, and ${\theta}_d$ converge at the saddle point while minimizing the model loss Eq.~(\ref{eq:minmax}) through the following minimax game:

\begin{equation}
    ({\hat{\theta}}_f, {\hat{\theta}}_c) = \argmin_{\theta_f, \theta_c}  L(\theta _{f}, \theta _{c}, \theta _{d}),
\end{equation}
\begin{equation}
    {\hat{\theta}}_d =  \argmax_{\theta_d} L(\theta _{f}, \theta _{c}, \theta _{d}).
\end{equation}

Although domain adversarial training shows promising results, this approach can lead to negative transfer between source classes and target classes in a PDA scenario.
For successful PDA, we need to isolate the source-private classes during the domain adaptation process.

\subsection{ Associative Partial Domain Adaptation}
\label{ssec:APDA}
To address negative transfer in PDA, we propose Associative Partial Domain Adaptation (APDA).
The overall flow of APDA can be seen in Fig.~\ref{fig:Main_pipeline}.
As in conventional domain adversarial networks, our model architecture consists of a feature extractor $F$, a label classifier $C$, and an adversarial domain discriminator $D$. 
Also, our model involves an auxiliary label classifier $C'$ and an auxiliary domain discriminator $D'$ to compute sample-level commonness $w(\mathbf{x}_{s})$ which indicates the probability of a source sample $\mathbf{x_s}$ belonging to the target label set $\mathbb{C}_{t}$.
More precisely, to obtain  $w(\mathbf{x}_{s})$, we use the discriminator logit $D'(x)$ and the normalized self-entropy ${E}(x) = -\frac{1}{\mathrm{log}N_c} \sum {l(x)} \mathrm{log}({l(x}))$ of samples where $l(x)$ denotes the predicted probability by classifier $C'$, and $N_c$ represents the total number of classes.

{
For clarity, we define $E_{Sp}, E_{Sc}$, and $E_{T}$ as the self-entropy value of source-private samples, source-common samples, and target samples, respectively.
Since $C'$ is trained only on the source data, the self-entropy of source samples should be lower than that of the target samples, \ie $E_{Sp}, E_{Sc} < E_{T}$.
Moreover, as source and target distributions are aligned from domain adversarial training, source-common samples have a higher probability of getting closer to the target samples than source-private samples.
As a result, the self-entropy of source-common samples is higher than that of source-private classes, i.e., $E_{Sp} < E_{Sc}.$
Overall, we can conclude that $E_{Sp} < E_{Sc} < E_{T}$.
For the  discriminator logit, the relationship across source-private, source-common, and target samples can be defined as $D'_{Sp} > D'_{Sc} > D'_{T}$ since the discriminator is trained to
predict samples from source (or target) domain as 1 (or 0).
}
Note that $D'$ is a two-class classifier called a domain discriminator, and $D$ is an adversarial domain discriminator with a gradient reversal layer \cite{ganin2015unsupervised}.
As a consequence, we formulate the sample-level label commonness:
\begin{equation}
\label{eq:weight}
    w(\mathbf{x}_{s}) =  {E}(C'(G(F(\mathbf{x}_{s}))))  - D'(G(F(\mathbf{x}_{s}))),
\end{equation}

Here, we subtract $D'(\cdot)$ from $E'(\cdot)$  and then normalize it into [0, 1], so $w(\mathbf{x}_{s})$ is always positive.
Importantly, sample-level weighting alone cannot handle anomaly samples, so we calculate commonness based on the features passed through the class relational graph $G(\cdot|\theta_{g})$ (See Fig.~\ref{fig:Main_pipeline}).
The details of class relational graph will be elaborated in the next subsection.
Once $w(\mathbf{x}_{s})$ is obtained at sample level, the label classifier $C$ and the adversarial domain discriminator $D$ are trained according to  $w(\mathbf{x}_{s})$ as follows: 
\begin{equation}
 \label{eq:lc}
    L_{c}=\mathbb{E}_{(\mathbf{x}_{s},y_s)\sim p_{s}}[w(\mathbf{x}_s) L_{y}(C(F(\mathbf{x}_{s})), y_s)], 
\end{equation}
\begin{equation}
\begin{split}
    L_{d} = & -\mathbb{E}_{(\mathbf{x}_{s})\sim p_{s}}[w(\mathbf{x}_s)logD(G(F(\mathbf{x}_{s})))]  \:\;\; \\
&-\mathbb{E}_{(\mathbf{x}_{t})\sim p_{t}}[log(1-D(G(F(\mathbf{x}_{t}))))].
\end{split}
\end{equation}
where $ L_{c}$ and $L_{d}$ are identical to conventional losses for label classifier and domain discriminator in standard domain adaptation except for the presence of $w(\mathbf{x_s})$.
{
Notice that the label classifier $C$ takes features directly from the feature extractor $F$, not from the graph module $G$ as shown in  Eq.~(\ref{eq:lc}) and the top of Fig \ref{fig:Main_pipeline}.
This architecture design effectively solves the issue where sufficient source and target samples are required to construct the graph even during the test phase, \ie model inference.

\textbf{Class Relational Graph:}
\label{sssec:crg}  
The main objective of Class Relational Graph (CRG) is to uncover the intrinsic structure by label-based feature propagation in both intra-domain and inter-domain.
Intra-domain propagation enables to select out non-trivial anomaly samples in each source-private class while inter-domain propagation encourages positive transfer by mapping nearby target samples and source samples with high label-commonness.
To propagate information between samples with high label association, we exploit graph convolutional networks.
Unlike standard convolutions, the goal of graph convolutional networks is to learn a function on a graph structure, which takes node features $H^l \in \mathbb{R}^{n\times d}$ and the corresponding adjacency matrix $A \in \mathbb{R}^{n \times n}$ as inputs, and updates the node features as $H^{l+1} \in \mathbb{R}^{n\times d'}$. Concretely, graph convolutional operation can be formulated as below:

\begin{equation}
 \label{eq:gcn}
\begin{split}
   {H}^{l+1} = \sigma({\widehat{A}}{H}^{l}{W}^{l}),
\end{split}
\end{equation}
\begin{equation}
 \label{eq:gcn2}
\begin{split}
  \widehat{A} = \widetilde{D}^{-\frac{1}{2}}\widetilde{{A}}\widetilde{D}^{-\frac{1}{2}},
\end{split}
\end{equation}
where  ${W}^{l}\in \mathbb{R}^{d\times d'}$ is a trainable weight matrix and  $\sigma(\cdot)$ denotes an activation function.
Note that  $\widetilde{A} = A + I $ is the adjacency matrix including self-connections and $\widetilde{D}_{ii} = \sum_{j}\widetilde{A}_{ij}$ is a degree matrix of $\widetilde{A}$.
Each element in the adjacency matrix (\ie edge) can be defined as hard or soft, depending on the situation.
As a result, graph convolutions can learn the intrinsic structure through iterative propagation between edge-connected nodes.

\begin{figure}[t]
\begin{center}
\def\arraystretch{0.5}
\begin{tabular}{@{}c@{\hskip 0.05\linewidth}c@{}c}
\includegraphics[width=0.4\linewidth]{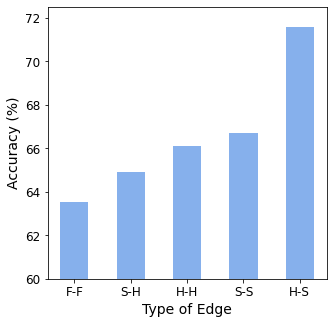} &
\includegraphics[width=0.41\linewidth]{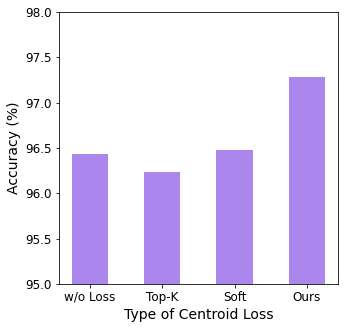} 
\\
{\hspace{2.7mm} (a) } & {\hspace{2.7mm}(b)}\\
\end{tabular}
\vspace*{-0.2in}
\end{center}
\caption{(a) Performance change by edge type on Office-Home.
``F-F'' denotes edge with feature similarity. The left side indicates the source label and the right side indicates the target label. ``H'' denotes hard label while ``S'' means soft label
(b) Performance change with respect to the configuration of the centroid-like loss on Office-31.
}
\label{fig:intermidiate_exp}
\end{figure}

The key challenge when applying the graph-based operation to non-structural inputs is the definition of nodes and edges. 
It is natural to use features from feature extractor as nodes in the proposed CRG.
However, edges can be defined in a wide variety of ways, resulting in very large performance changes.
Previous work~\cite{ma2019gcan} use similarities between features as edges~\cite{ma2019gcan}.
However, this approach cannot utilize class information that is considered important in domain adaptation.
Instead, we compute the edge based on the class relationship by using hard labels from the source domain and soft labels from the target domain.
For this, we  define different class information according to the domain type:
\begin{equation}
\mathbf{\widetilde{y}} =\left\{\begin{matrix} &\mathbf{y},  \hspace{16mm}  \textup{if} \hspace{2mm} \mathbf{x} \in D_{s}
\\ 
&C(F(\boldsymbol{\mathbf{x}})), \hspace{5mm} \textup{if}  \hspace{2mm}  \mathbf{x} \in D_{t}
\end{matrix}\right.,
\end{equation}
where we use ground truth labels in the source domain, specifically one-hot vectors. Since target labels are not given in PDA, we exploit soft-labels, predicted class probability from the source classifier $C$. 
Finally, we can generate an class-based adjacency matrix as $ \widetilde{A}_{ij} = \mathbf{\widetilde{y}}_{i}^{T}\mathbf{\widetilde{y}}_{j}$.

We can see several meaningful observations with respect to the edge types from Fig. \ref{fig:intermidiate_exp} (a).
Using probabilistic (soft) labels as class information for nodes in the source domain results in performance degradation.
It is no surprise that ground truth labels provide more accurate information than the classifier's probability information in the source domain.
In contrast, using probabilistic labels in the target domain outperforms using hard labels with the highest probability. 
Surprisingly, our proposed class-based edges show better performance with higher margins than feature similarity-based edges, regardless of the type.
}

For more sophisticated CRG analysis, we can divide our label-based message passing into intra-domain and inter-domain.
1) Intra-domain: nodes of the same class in the source domain are connected to hard edges; so our CRG can reveal the intrinsic class-wise structure.
On the other hand, the edges of the target domain are generated from predicted probabilistic labels, resulting in a soft link between the target samples.
2) Inter-domain: classifying source samples through a label classifier results in high probabilities for the common classes and low probabilities for the source-private classes as discussed in \cite{cao2018partialtransfer,cao2018partial}. From this, we can expect that source samples connected to the target samples with strong edges are likely to be of the common classes.  
Consequently, samples connected by edges have similar values through message passing, thus encouraging positive transfer.

\textbf{Confidence-guided Loss:}
\label{sssec:cgl} 
We can expect that classes with high average commonness are more likely to belong to a common label set, otherwise they are likely to be source-private classes.
To fully utilize class-level commonness,  we propose a novel confidence-guided loss.
The main objective of this guidance loss is to keep the source-private samples away from the target samples while encouraging the common classes closer to the target samples.
More precisely, we encourage the centroid of the source samples with a high class-level commonness and the target samples with the corresponding pseudo labels closer in the feature space.
We average sample-level commonness obtained from  Eq. \ref{eq:weight}) for each class to obtain class-level commonness.
Furthermore, our confidence-guided loss is specially designed so that the centroid of the source samples with a low commonness is far away from the centroid of target samples with the same pseudo label.
To reduce errors caused by corner cases,  we apply confidence-guided loss only to top-$K $ classes and bottom-$K$ classes according to the value of class-level commonness: 
\begin{equation}
L_{cg} = \frac{1}{2K}(\sum _{i \in T_K} \left \| R^{i}_{s}-R^{i}_{t} \right \|^{2}_{2} \hspace{2mm}
-   \sum _{j \in B_K}   \left \| R^{j}_{s}-R^{j}_{t} \right \|^{2}_{2} ).
\label{eq:class-centroid}
\end{equation}
Here, $T_K$ and $B_K$ are top-$K$ class set and bottom-$K$ class set respectively. 
$R^i$ represents the centroid feature for $i$-th class obtained from the CRG features, \ie $G(F(\mathbf{x}))$.
Note that ground truth labels are used in the source domain while pseudo labels are used in the target domain.
If the total number of labels is larger than the batch size, categorical information in each batch is very sparse. This makes the model unstable and can lead to performance degradation.
Therefore, we empirically resolve this issue by applying exponential moving average over iteration steps.

Unlike conventional class-centroid loss \cite{xie2018learning} that always causes the distance between samples to be closer, to our best knowledge, this is the first guidance loss that makes the unrelated samples away from each other (see a negative sign in Eq. \ref{eq:class-centroid}). 
To show the effectiveness of repulsive forces in the proposed method, we compare our final model with three configurations as shown in Fig. \ref{fig:intermidiate_exp} (b).
{
In the figure,  
``w/o loss'' means our model without confidence-guided loss, \ie a
model using only CRG. 
``Top-K" denotes our model with loss from only top-K classes (\ie the first term in Eq. \ref{eq:class-centroid})
and ``Soft" represents our model with soft-weighting rather than our hard top-k thresholding.
``Ours'' refers to our final model including CRG and confidence-guided loss.
}
From the figure, we can see that learning with repulsive forces enhance positive transfer and suppress negative transfer effectively.

\begin{table*}[]
    \addtolength{\tabcolsep}{1.5pt}
    \centering
    \caption{Classification Accuracy (\%) on {Office-31}, {VisDA2017}, and {ImageNet-Caltech} ({ResNet-50})}
    \label{table:accuracy_officeic}
    \resizebox{1\textwidth}{!}{%
    \begin{tabular}{lccccccccccccccc}
        \toprule
        \multirow{2}{30pt}{\centering Method}\:\:\:\:\:\:\: &  \multicolumn{7}{c}{Office-31} && \multicolumn{3}{c}{VisDA2017} && \multicolumn{3}{c}{ImageNet-Caltech} \\
        \cmidrule{2-8} \cmidrule{10-12} \cmidrule{14-16}
        & \:A$\Rightarrow$W\: & \:D$\Rightarrow$W\: & \:W$\Rightarrow$D\: & \:A$\Rightarrow$D\: & \:D$\Rightarrow$A\: & \:W$\Rightarrow$A\: & \:Avg\: && \: R$\Rightarrow$S \: & \: S$\Rightarrow$R \:& \:Avg\: &&  \:I$\Rightarrow$C\: & \: C$\Rightarrow$I\: & \:Avg\:  \\
        \midrule
        ResNet~\cite{he2016deep}& 75.59 & 96.27 & 98.09 & 83.44 & 83.92 & 84.97 & 87.05 &&  64.28 & 45.26 & 54.77 && 69.69 & 71.29 & 70.49 \\
        DAN~\cite{long2015learning} & 59.32 & 73.90& 90.45 & 61.78 & 74.95 & 67.64 & 71.34 && 68.35 & 47.60 & 57.98  && 71.30 & 60.13 & 65.72 \\
        DANN~\cite{ganin2015unsupervised} & 73.56 & 96.27 & 98.73 & 81.53 & 82.78 & 86.12 & 86.50 && 73.84 & 51.01 & 62.43  && 70.80 & 67.71 & 69.23 \\
        RTN~\cite{long2016unsupervised} & 78.98 & 93.22 & 85.35 & 77.07 & 89.25 & 89.46 & 85.56 && 72.93 & 50.04 & 61.49 && 75.50 & 66.21 & 70.85 \\
        IWAN~\cite{zhang2018importance} & 89.15 & 99.32 & 99.36 & 90.45 & 95.62 & 94.26 & 94.69 &&  71.30 & 48.60 & 59.95  && 78.06 & 73.33 & 75.70 \\
        SAN~\cite{cao2018partialtransfer} & 93.90 & 99.32 & 99.36 & 94.27 & 94.15 & 88.73 & 94.96 && 69.70 & {49.90} & 59.80 && 77.75 & 75.26 & 76.51 \\
        PADA~\cite{cao2018partial} &86.54 & 99.32 & {100.0} & 82.17 & 92.69 & {95.41} & 92.69 && 76.50 & 53.50 & 65.00  && 75.03 & 70.48 & 72.76\\
        DRCN~\cite{li2020deep} & {90.80} & {100.0} & {100.0} & {86.00} & {95.60} & \textbf{95.80} & {94.30} &&  {73.20} & 58.20 & 65.70  && 75.30 & 78.90 & 71.40 \\
        ETN~\cite{cao2019learning} & {94.52} & {100.0} & {100.0} & {95.03} & \textbf{96.21} & 94.64 & {96.73} &&  - & - & -  && \textbf{83.23} & 74.93 & 79.08 \\
        \midrule
        \textrm{Base} (\textrm{ours}) & 87.79 & 100.0 & 100.0 & 93.63 & 93.94 & 95.34 & 95.11 &&  71.89 & 59.30 & 65.59  && 72.03 & 72.48 & 72.25 \\
        \textrm{CRG} (\textrm{ours}) & 93.22 & 100.0 & 100.0 & 94.90 & 94.88 & 95.52 & 96.42 && 79.82 & 66.80 & 73.31  && 77.20 & 79.57 & 78.83 \\
        {APDA} (ours) & \textbf{96.61} & \textbf{100.0} & \textbf{100.0} & \textbf{96.17} & 95.50 & 95.40 & \textbf{97.28} &&  \textbf{80.66} & \textbf{67.68} & \textbf{74.17} && {79.10} & \textbf{80.17} & \textbf{79.64} \\
        \bottomrule
    \end{tabular}%
    }
\end{table*}

\begin{table*}[]
    \addtolength{\tabcolsep}{1.5pt}
    \centering
    \caption{Classification Accuracy (\%)  on {Office-Home}  ({ResNet-50})}
    \label{table:accuracy_officehome}
    \resizebox{\textwidth}{!}{%
    \begin{tabular}{lccccccccccccc}
        \toprule
        \multirow{2}{30pt}{\centering Method}\:\:\:\:\:\:\: & \multicolumn{13}{c}{Office-Home} \\
        \cmidrule{2-14}
        & {Ar}$\Rightarrow${Cl} & {Ar}$\Rightarrow${Pr} & {Ar}$\Rightarrow${Rw} & {Cl}$\Rightarrow${Ar} & {Cl}$\Rightarrow${Pr} & {Cl}$\Rightarrow${Rw} & {Pr}$\Rightarrow${Ar} & {Pr}$\Rightarrow${Cl} & {Pr}$\Rightarrow${Rw} & {Rw}$\Rightarrow${Ar} & {Rw}$\Rightarrow${Cl} & {Rw}$\Rightarrow${Pr} & \:\:Avg\:\: \\
        \midrule
        ResNet~\cite{he2016deep} & 46.33 & 67.51 & 75.87 & 59.14 & 59.94 & 62.73 & 58.22 & 41.79 & 74.88 & 67.40 & 48.18 & 74.17 & 61.35 \\
        DANN~\cite{ganin2015unsupervised} & 43.76 & 67.90 & 77.47 & 63.73 & 58.99 & 67.59 & 56.84 & 37.07 & 76.37 & 69.15 & 44.30 & 77.48 & 61.72 \\
        ADDA~\cite{tzeng2017adversarial} & 45.23 & 68.79 & 79.21 & 64.56 & 60.01 & 68.29 & 57.56 & 38.89 & 77.45 & 70.28 & 45.23 & 78.32 & 62.82 \\
        RTN~\cite{long2016unsupervised} & 49.31 & 57.70 & 80.07 & 63.54 & 63.47 & 73.38 & 65.11 & 41.73 & 75.32 & 63.18 & 43.57 & 80.50 & 63.07 \\
        IWAN~\cite{zhang2018importance} & 53.94 & 54.45 & 78.12 & 61.31 & 47.95 & 63.32 & 54.17 & 52.02 & 81.28 & {76.46} & 56.75 & 82.90 & 63.56 \\
        SAN~\cite{cao2018partialtransfer} & 44.42 & 68.68 & 74.60 & {67.49} & 64.99 & {77.80} & 59.78 & 44.72 & 80.07 & 72.18 & 50.21 & 78.66 & 65.30 \\
        PADA~\cite{cao2018partial} &51.95 & 67.00 & 78.74 & 52.16 & 53.78 & 59.03 & 52.61 & 43.22 & 78.79 & 73.73 & 56.60 & 77.09 & 62.06 \\
        DRCN~\cite{li2020deep} & {54.00} &{76.40} & 83.00 & 62.10 & {64.50} & 71.00 & \textbf{70.80} & {49.80} & {80.50} & 77.50 & \textbf{59.10} & {79.90} & {69.00} \\
        ETN~\cite{cao2019learning} & \textbf{59.24} & {77.03} & 79.54 & 62.92 & {65.73} & 75.01 & {68.29} & \textbf{55.37} & \textbf{84.37} & 75.72 & {57.66} & \textbf{84.54} & {70.45} \\
        \midrule
        {Base} (ours) & 53.07 & 70.25 & 83.99 & 63.91 & 62.75 & 74.65 & 62.72 & 47.04 & 81.17 & 77.04 & 53.13 & 81.90 & 67.64 \\
        {CRG} (ours) & 53.61 & 74.96 & {84.70} & 73.09 & 69.08 & 79.79 & 67.86 & 50.57 & 79.85 & {78.05} & {56.06} & 82.52 & 70.85 \\
        {APDA} (ours) & {54.39} & \textbf{77.98} & \textbf{85.26} & \textbf{73.92} & \textbf{71.60} & \textbf{82.72} & {69.61} & {50.87} & {81.83} & \textbf{78.15} & 55.70 & {82.58} & \textbf{72.05} \\
        \bottomrule
    \end{tabular}%
    }
\end{table*}

\section{Experiments}
\label{sec:exp}

\subsection{Experimental Setup}
We perform evaluations on four public datasets: Office-31~\cite{saenko2010adapting}, Office-Home~\cite{venkateswara2017deep}, VisDA2017~\cite{peng2017visda} and ImageNet-Caltech~\cite{russakovsky2015imagenet}.
Across all datasets, we follow the official partial domain adaptation protocol for comparison with existing approaches.

\textbf{Office-31} is a benchmark dataset from three different domains and contains 4,652 images and 31 categories.
We use all 31 categories for the source domain and 10 categories shared between office-31 and Caltech-256 for the target domain.
\textbf{Office-Home} consists of 65 categories and is collected from four different domains.
We use all 65 categories for the source domain and use the first 25 categories in alphabetical order for the target domain.
\textbf{VisDA2017} is designed for a synthetic-to-real visual domain shift scenario. This large-scale dataset contains 152,397 synthetic (S) images and 55,388 real-world (R) images.
All 12 categories are used as source domain classes, whereas only the first 6 categories in alphabetical order are used for the target domain.
\textbf{ImageNet-Caltech} is also a large-scale dataset considering practical scenarios.
This setting utilizes ImageNet-1K (I) and Caltech-256 (C) to configure two partial domain adaptation scenarios. 
Two datasets share 84 common classes, so we perform the experiment on two scenarios: ImageNet-1K $\Rightarrow$ Caltech-256 and Caltech-256 $\Rightarrow$ ImageNet84.
When ImageNet is used as the target domain, the validation set of ImageNet is used to prevent the influence of our model trained in the training set.

For a fair comparison of the proposed method with the state-of-the-art methods, we use ResNet-50 \cite{he2016deep} and VGG-16 \cite{simonyan2014very} pre-trained from ImageNet as the base models.
The base learning rate is set to 0.001 and all the fine-tuned layers are optimized with a learning rate of 0.01.
We adjust the learning rate  using $lr_p = \frac{lr_0}{{(1+\alpha p)}^\beta}$, where $lr_0$ is a base learning rate, $p$ is a relative step  that changes from 0 to 1  as the training progresses, $\alpha$ = 10, and $\beta$ = 0.75.
The adaptation factor $\lambda$ is gradually adjusted from 0 to 1 during the training process, taking into account the unstable domain discriminator in the early training stages.

\subsection{Experimental Results}
As can be seen in Table \ref{table:accuracy_officeic} and Table \ref{table:accuracy_officehome}, we compare the proposed method with existing methods: 
ResNet-50 \cite{he2016deep}, DAN \cite{long2015learning}, DANN \cite{ganin2015unsupervised}, RTN \cite{long2016unsupervised}, ADDA \cite{tzeng2017adversarial},  SAN \cite{cao2018partialtransfer},  PADA \cite{cao2018partial}, IWAN \cite{zhang2018importance}, DRCN \cite{li2020deep} and ETN \cite{cao2019learning}.
We cite the results reported by other publications when the experimental setup is the same. 
When we need to perform experiments that are not reported in the previous works, we compare them
using the official source codes under the same experimental protocols.
The ``Base'' reported in the tables denotes the variation of APDA 
which does not utilize the graph module.
In other words, the feature after $F$ is directly passed into $C, D, C'$, and $D'$.
``CRG" stands for our model that only considers class relational graphs.
``APDA" refers to our final model with confidence-guided loss.

From the tables, we can see the following observations.
 Standard DA methods (\eg DAN, DANN, and RTN) show the lower performance than using a basic ResNet in several adaptation scenarios, which indicates that directly domain alignment without considering source-private classes induces negative transfer.  
Previous PDA methods (\eg SAN, PADA, IWAN, DRCN, and ETN) outperform ResNet and other standard DA methods, demonstrating that class-level or sample-level weighting mechanisms effectively mitigate negative transfer.
Nevertheless, existing approaches are still suffering from negative transfer by non-trivial anomaly samples within source-private classes.
Through this paper, we assert that PDA models should fully utilize intra- and inter-domain class relationships to address the above issue.
As a result, APDA outperforms state-of-the-art methods across all datasets, which indicates that it is important to handle anomaly samples in the PDA scenario.

There are relatively few performance differences between the methods in Office-31 due to the high representation power of ResNet.
Therefore, we perform a comparison experiment using VGG-16 \cite{simonyan2014very} as the base architecture to compare the contribution of each approach to solving PDA.
From Table \ref{table:accuracy_vggoffice}, we can see the superiority of the proposed method is consistent with the existing results.

\begin{table*}[]
    \addtolength{\tabcolsep}{2.5pt}
    \centering
    \caption{Classification Accuracy (\%)  on {Office-31} ({VGG-16})}
    \label{table:accuracy_vggoffice}
    \resizebox{0.76\textwidth}{!}{%
    \begin{tabular}{lccccccc}
        \toprule
        \multirow{2}{30pt}{\centering Method}\:\:\:\:\:\:\:  &  \multicolumn{7}{c}{Office-31} \\
        \cmidrule{2-8}
        & \:\:\:A$\Rightarrow$W\:\:\: & \:\:\:D$\Rightarrow$W\:\:\: & \:\:\:W$\Rightarrow$D\:\:\: & \:\:\:\:\:A$\Rightarrow$D\:\:\: & \:\:\:D$\Rightarrow$A\:\:\: & \:\:\:W$\Rightarrow$A\:\:\: & \:\:\:\:Avg\:\:\:\: \\
        \midrule
        VGG~\cite{he2016deep}& 60.34 & 97.97 & 99.36 & 76.43 & 72.96 & 79.12 & 81.03 \\
        DAN~\cite{long2015learning} & 58.78 & 85.86 & 92.78 & 54.76 & 55.42 & 67.29 & 69.15  \\
        DANN~\cite{ganin2015unsupervised} & 50.85 & 95.23 & 94.27 & 57.96 & 51.77 & 62.32 & 68.73 \\ ADDA~\cite{ganin2015unsupervised} & 53.28 & 94.33 & 95.36 & 58.78 & 50.24 & 63.34 & 69.22 \\
        RTN~\cite{long2016unsupervised} & 69.35 & 98.42 & 99.59 & 75.43 & 81.45 & 82.98 & 84.54  \\
        IWAN~\cite{zhang2018importance} & 82.90 & 79.75 & 88.53 & 90.95 & 89.57 & 93.36 & 87.51 \\
        SAN~\cite{cao2018partialtransfer} & 83.39 & 99.32 & 100.0 & 90.70 & 87.16 & 91.85 & 92.07  \\
        PADA~\cite{cao2018partial} &86.05& 99.42 & {100.0} & 81.73 & 93.00 & \textbf{95.26} & 92.54 \\
        ETN~\cite{cao2019learning} & {85.66} & {100.0} & {100.0} & {89.43} & \textbf{95.93} & 92.28 & {93.88} \\
        \midrule
        \textrm{Base} (\textrm{ours}) & 88.81 & \textbf{100.0} & 99.36 & 84.32 & 93.73 & 94.26 & 93.41\\
        \textrm{CRG} (\textrm{ours}) & 90.16 & 99.32 & 100.0 & 86.62 & 93.94 & 94.46 & 94.08 \\
        {APDA} (ours) & \textbf{91.18} & {99.66} & \textbf{100.0} & \textbf{92.35} & 94.05 & 94.25 & \textbf{95.24} \\
        \bottomrule
    \end{tabular}%
    }
\end{table*}

\begin{figure}[]
\begin{center}
\def\arraystretch{0.5}
\begin{tabular}{@{}c@{\hskip 0.01\linewidth}c@{\hskip 0.01\linewidth}c@{\hskip 0.01\linewidth}c@{\hskip 0.01\linewidth}c@{\hskip 0.01\linewidth}c}
\includegraphics[width=0.48\linewidth]{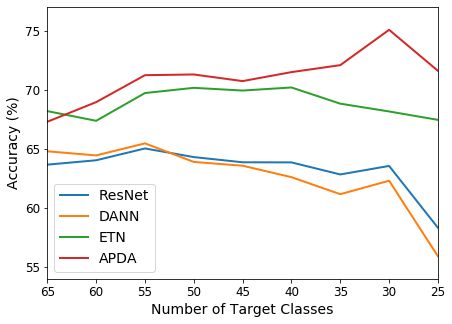}  &
\includegraphics[width=0.49\linewidth]{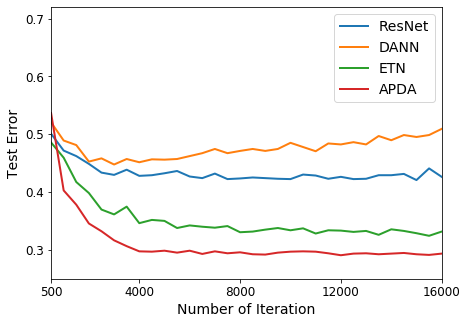} 
\\
{\hspace{4.3mm} (a) } & {\hspace{4.6mm}(b) }\\
\end{tabular}
\vspace*{-0.2in}
\end{center}
\caption{
(a) Test accuracy with respect to the number of target classes.  (b) Test error of target samples with regard to training iteration.
}
\label{fig:cl_pr}
\end{figure}

\begin{figure}[]
\begin{center}
\def\arraystretch{0.5}
\begin{tabular}{@{}c@{\hskip 0.01\linewidth}c@{\hskip 0.01\linewidth}c@{\hskip 0.01\linewidth}c@{\hskip 0.01\linewidth}c@{\hskip 0.01\linewidth}c}
\includegraphics[width=0.32\linewidth]{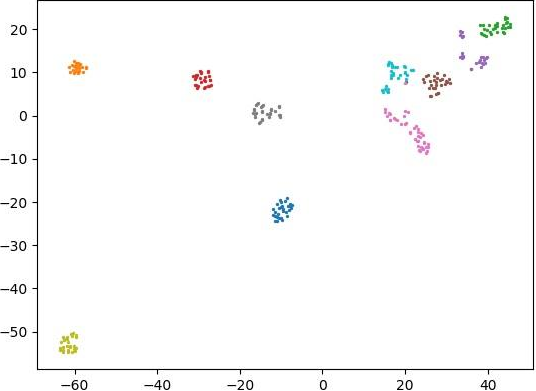} &
\includegraphics[width=0.32\linewidth]{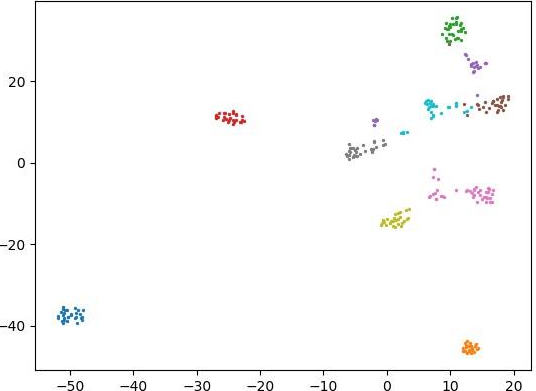}  &
\includegraphics[width=0.32\linewidth]{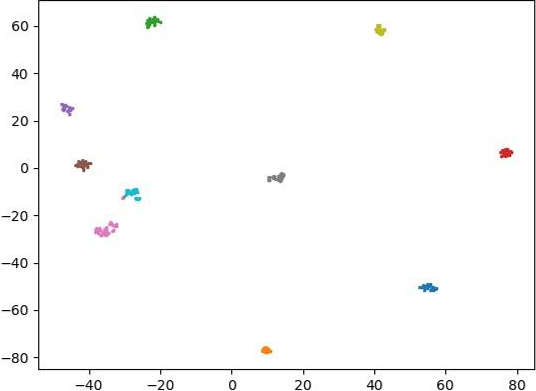} 
\\
{\hspace{2mm} (a) DANN } & {\hspace{2mm}(b) PADA } &  {\hspace{3mm}(c) APDA (ours) }  \\
\end{tabular}
\vspace*{-0.2in}
\end{center}
\caption{
Visualization of feature space. Each color represents different class.}
\vspace*{-0.15in}
\label{fig:tsne}
\end{figure}

\begin{figure}[]
\begin{center}
\def\arraystretch{0.5}
\begin{tabular}{@{}c@{\hskip 0.05\linewidth}c@{}c}
\includegraphics[width=0.49\linewidth]{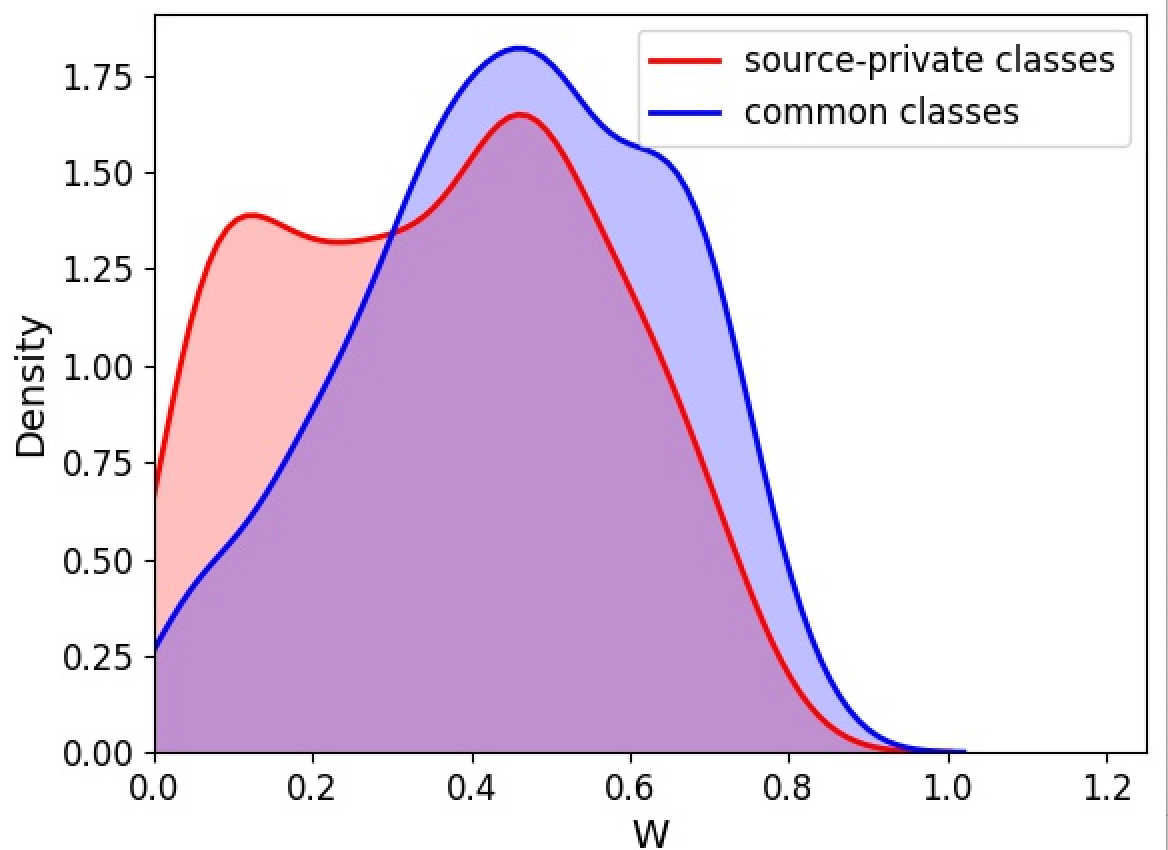} &
\includegraphics[width=0.49\linewidth]{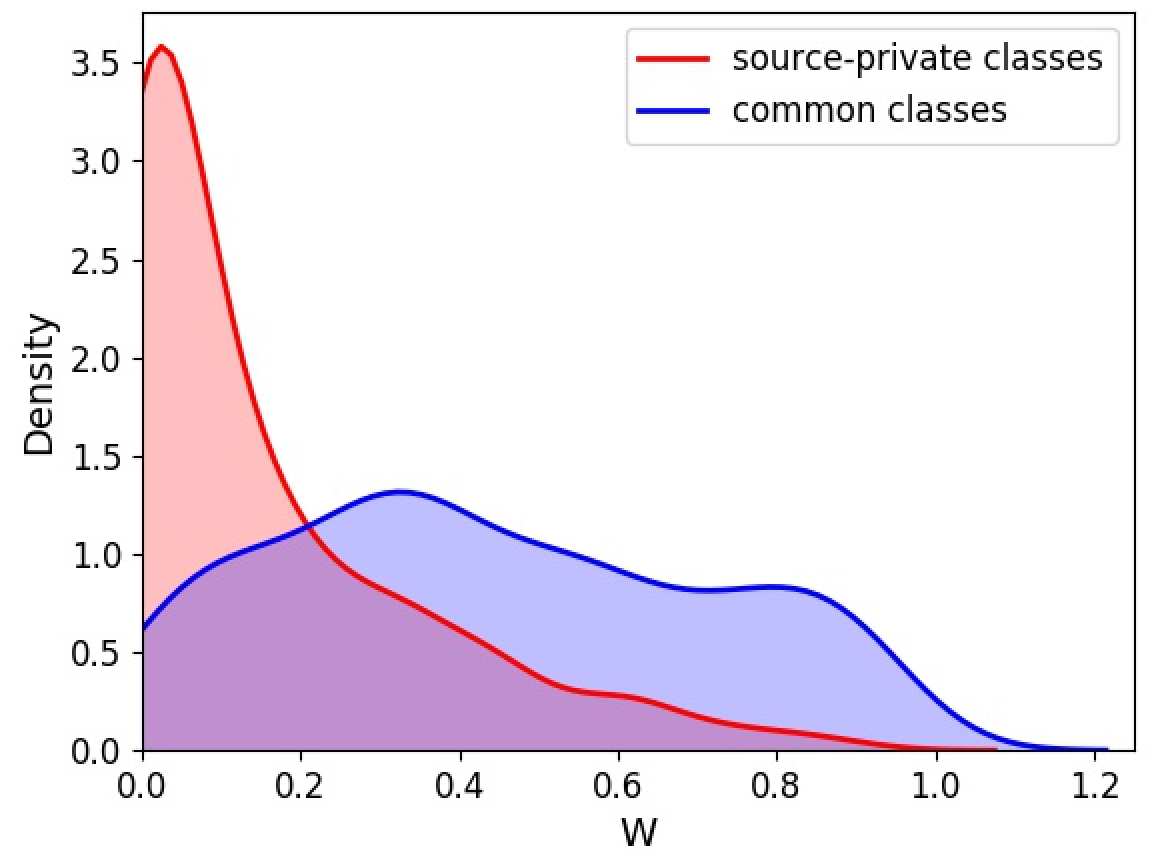} 
\\
{\hspace{2.7mm} (a) ETN } & {\hspace{2.7mm}(b)  APDA (ours)}\\
\end{tabular}
\vspace*{-0.2in}
\end{center}
\caption{Density function of the commonness score.
}
\label{fig:wdensity}
\end{figure}

\subsection{Empirical Analysis}
\label{ssec:Analysis}
To quantitatively and qualitatively demonstrate how our associative modeling efficiently solves PDA problems, we present extensive ablation studies.

\textbf{Class Overlap:}
\label{sssec:class_overlap}
We evaluate the performance change according to the number of target classes on {Cl} $\Rightarrow$ {Pr}  in Office-Home.
From Fig. \ref{fig:cl_pr} (a), we can see that ResNet achieves better performance than DANN when the number of target classes is smaller than 50, which indicates the negative transfer in standard DA.
ETN shows robust performance, but it still shows performance degradation.
Our APDA shows its effectiveness even at the small size of the target classes.

\textbf{Convergence Performance:}
\label{sssec:test_error}
We compare the convergence of PDA approaches by investigating the test error on {Cl} $\Rightarrow$ {Pr}  in Office-Home.
From  Fig. \ref{fig:cl_pr} (b), we can infer that standard DA methods steadily undergo performance degradation due to negative transfer. 
Unlike these approaches, our proposed model converges quickly and stably.

\textbf{Feature Visualization:} 
\label{sssec:t_sne}
To visually examine the effectiveness of the proposed method, we use t-SNE \cite{maaten2008visualizing} embedding  on {A} $\Rightarrow$ {W}  in Office-31. 
We can clearly see that features learned by DANN and PADA are not clustered as clearly as ours as shown in Fig. \ref{fig:tsne}.
This indicates that our model successfully mitigates negative transfer and consequently leads to discriminative feature space.

\textbf{Weight Visualization:}
\label{sssec:w_visual}
Figure \ref{fig:wdensity} shows the approximate density function of sample-level commonness on {Cl} $\Rightarrow$ {Ar} in Office-Home.
Both approaches show less weight for the source-private class (red) and show higher weights for the common class (blue).
However, our method involves more weight with almost zero values for source-private classes.

{
\subsection{Parameter Sensitivity}
\label{sssec:param_sensitivity}
We present parameter sensitivity on {Cl} $\Rightarrow$ {Pr} in Office-Home.
}

\begin{figure}[]
\vspace*{-0.1in}
\begin{center}
\def\arraystretch{0.5}
\begin{tabular}{@{}c@{\hskip 0.01\linewidth}c@{\hskip 0.01\linewidth}c@{\hskip 0.01\linewidth}c@{\hskip 0.01\linewidth}c@{\hskip 0.01\linewidth}c}
\includegraphics[width=0.33\linewidth]{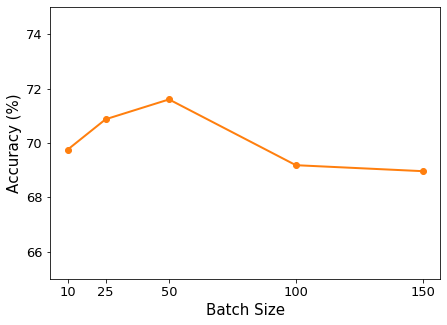} &
\includegraphics[width=0.33\linewidth]{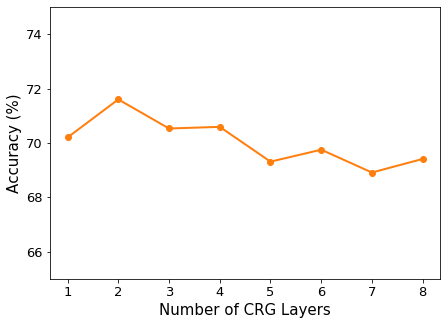}  &
\includegraphics[width=0.33\linewidth]{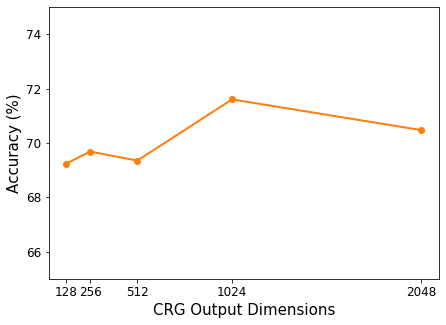} 
\\
{\hspace{4.5mm} (a)  } & {\hspace{4.5mm}(b)  } &  {\hspace{4.5mm}(c)  }  \\
\end{tabular}
\vspace*{-0.2in}
\end{center}
\caption{
Parameter sensitivity analysis on a CRG module. 
(a) Batch size.
(b) The number of graph convolutional layers.
(c) Output dimensions of the CRG. }
\vspace*{-0.15in}
\label{fig:CRG_analysis}
\end{figure}

\textbf{Class Relational Graph (Batch Size):}
Batch size is important in our approach since the CRG module constructs a graph structure from samples in each batch.
Fig. \ref{fig:CRG_analysis} (a) show the robustness of our ADPA with respect to the batch size.

\begin{figure}[]
\vspace*{-0.1in}
\begin{center}
\def\arraystretch{0.5}
\begin{tabular}{@{}c@{\hskip 0.01\linewidth}c@{\hskip 0.01\linewidth}c@{\hskip 0.01\linewidth}c@{\hskip 0.01\linewidth}c@{\hskip 0.01\linewidth}c}
\includegraphics[width=0.33\linewidth]{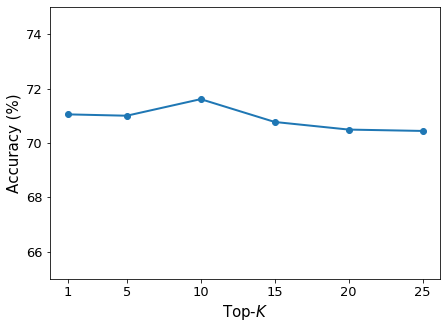} &
\includegraphics[width=0.33\linewidth]{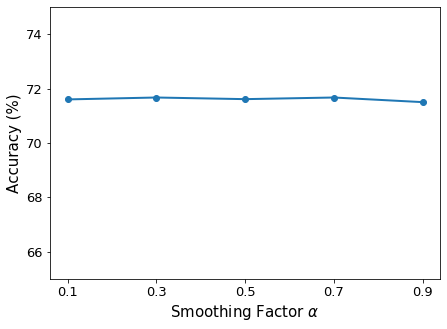}  &
\includegraphics[width=0.33\linewidth]{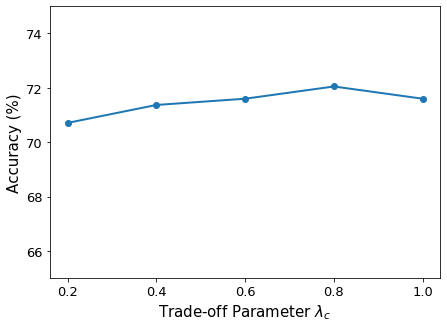} 
\\
{\hspace{4.5mm} (a)  } & {\hspace{4.5mm}(b)  } &  {\hspace{4.5mm}(c)  }  \\
\end{tabular}
\vspace*{-0.2in}
\end{center}
\caption{
Parameter sensitivity analysis related to confidence-guided loss. 
(a) Different $K$ in Top-K.
(b) Smoothing factor $\alpha$ in moving average. 
(c) Trade-off parameter $\lambda_{c}$.}
\vspace*{-0.15in}
\label{fig:Cgloss_analysis}
\end{figure}

\textbf{Class Relational Graph (Number of Layers):}
We compare the performance by changing the number of layers from $1$ to $8$. 
From Fig. \ref{fig:CRG_analysis} (b), we observe that our APDA achieves promising results when the number of layers is less than 5.
As the layers go deeper, there is performance degradation due to the over-smoothing effect.

\textbf{Class Relational Graph (Output Dimensions):}
We analyze the representation power of the CRG module by changing its output feature dimension from $128$ to $2048$.
Figure \ref{fig:CRG_analysis} (c) shows that higher feature dimensions (\ie 1024 and 2048) lead to significant performance improvements.

\textbf{Confidence-guided Loss (Top-$K$):}
We apply confidence-guided loss only to top-$K$ classes and bottom-$K$ classes. We design experiments to show the robustness of our method by varying the $K$ value. 
As shown in Fig. \ref{fig:Cgloss_analysis} (a), APDA shows consistently reasonable performance.

\textbf{Confidence-guided Loss (Moving Average):}
 To resolve the issue from sparse categorical information in each batch,  we apply exponential moving average over iteration steps:
    ${W^{t}_c} \leftarrow \alpha{{W}^{t}_c} + (1-\alpha){W^{t-1}_c}$.
We adopt different $\alpha$ values (\ie [0,1]) to evaluate the performance.  
From Fig. \ref{fig:Cgloss_analysis} (b), we conclude that our confidence-guided loss is robust to the smoothing factor $\alpha$.

\textbf{Confidence-guided Loss (Trade-off Parameter):}
We set a trade-off parameter $\lambda_{c}$ to control the balance between the confidence-guided loss and other losses, \eg cross-entropy loss and domain adversarial loss.
From Fig. \ref{fig:Cgloss_analysis} (c), we can see that our approach yields better results when  $\lambda_{c} \geq 0.6$.

\section{Conclusion}
In this paper, we have proposed a novel partial domain adaptation scheme called  Associative Partial Domain Adaptation (APDA).
Our key idea is to handle the weight of a domain confusion loss and a classification loss based on the associated information between source and target samples.
Concretely, we discover intra-domain association for reducing the negative transfer effect caused by sample variation within one class. 
Simultaneously, inter-domain association enhances positive transfer by mapping nearby target and source samples with high label-commonness. 
We have demonstrated the efficiency and effectiveness of our  APDA on various benchmarks and achieved state-of-the-art performance consistently.

\bibliography{ref}

\end{document}